\documentclass[conference]{IEEEtran}
\IEEEoverridecommandlockouts
\usepackage{graphicx}
\usepackage{amsmath}
\usepackage{url}
\usepackage{rotating}
\usepackage{adjustbox}
\usepackage{booktabs}
\usepackage{hyperref}
\usepackage{multirow}
\usepackage{xcolor}

\ifCLASSINFOpdf
\else
\fi
\usepackage{tikz}
\usepackage{textcomp}
\usepackage{hyperref}
\usepackage{lipsum}

\newcommand\copyrighttext{%
  \footnotesize \textcopyright 2021 IEEE. Personal use of this material is permitted.
  Permission from IEEE must be obtained for all other uses, in any current or future
  media, including reprinting/republishing this material for advertising or promotional
  purposes, creating new collective works, for resale or redistribution to servers or
  lists, or reuse of any copyrighted component of this work in other works.
  DOI: \href{https://doi.org/10.1109/DSAA53316.2021.9564238}{10.1109/DSAA53316.2021.9564238}}
\newcommand\copyrightnotice{%
\begin{tikzpicture}[remember picture,overlay]
\node[anchor=south,yshift=10pt] at (current page.south) {\fbox{\parbox{\dimexpr\textwidth-\fboxsep-\fboxrule\relax}{\copyrighttext}}};
\end{tikzpicture}%
}

\begin{document}
%
\title{Improving Portuguese Semantic Role Labeling with Transformers and Transfer Learning
}

\author{\IEEEauthorblockN{Sofia Oliveira}
\IEEEauthorblockA{LIAAD - INESC TEC\\
Dept. of Computer Science - FCUP\\
University of Porto\\
Porto, Portugal\\
anasofia.oliveira@protonmail.com}
\and
\IEEEauthorblockN{Daniel Loureiro}
\IEEEauthorblockA{LIAAD - INESC TEC\\
Dept. of Computer Science - FCUP\\
University of Porto\\
Porto, Portugal\\
daniel.b.loureiro@inesctec.pt}
\and
\IEEEauthorblockN{Alípio Jorge}
\IEEEauthorblockA{LIAAD - INESC TEC\\
Dept. of Computer Science - FCUP\\
University of Porto\\
Porto, Portugal\\
amjorge@fc.up.pt
}}


%


\maketitle
\copyrightnotice
\begin{abstract}
The Natural Language Processing task of determining ``Who did what to whom" is called Semantic Role Labeling. For English, recent methods based on Transformer models have allowed for major improvements in this task over the previous state of the art. However, for low resource languages, like Portuguese, currently available semantic role labeling models are hindered by scarce training data. In this paper, we explore a model architecture with only a pre-trained Transformer-based model, a linear layer, softmax and Viterbi decoding. We substantially improve the state-of-the-art performance in Portuguese by over 15 F1. Additionally, we improve semantic role labeling results in Portuguese corpora by exploiting cross-lingual transfer learning using multilingual pre-trained models, and transfer learning from dependency parsing in Portuguese, evaluating the various proposed approaches empirically.

\end{abstract}



%
\IEEEpeerreviewmaketitle

\section{Introduction}

Semantic Role Labeling (SRL) is the task of determining ``Who did What to Whom, How, When and Where", i.e., identifying events in sentences, their participants and properties \cite{book,special}. The predicate (usually a verb) corresponds to the event and the participants are expressed by arguments, which take on a semantic role in relation to a predicate. SRL is commonly viewed as an intermediate task that can be useful for applications such as: question answering \cite{qa,qa2}; information extraction \cite{ie,ie2} and summarization \cite{summ,summ2}.

The most prominent Transformer-based Language Models (e.g. BERT \cite{bert}) are trained on English text, and have been a key component of state-of-the-art solutions for English SRL \cite{sota,structured}. For Portuguese, however, there has not been much work done, and the current state of the art, Falci et al. \cite{falci} is based on a model by He et al. \cite{he} that has since been surpassed in English.
Our work aims at transferring state-of-the-art techniques proposed for the English language to a lower resource language: Portuguese. 
We apply a monolingual model, BERTimbau \cite{bertpt}, to improve the current state of the art in Portuguese SRL.

Additionally, we investigate further approaches for reaching better performance in this task for Portuguese, namely by using multilingual models, using cross-lingual transfer learning and transfer learning from a model fine-tuned in a syntax-related task. We use syntax in our transfer learning approach due to its known close relation with semantics \cite{gildea_and_jurafsky}; in fact, SRL models have traditionally relied on syntax.

Our main contributions for the state of the art on the SRL task are the following.
\begin{itemize}
    \item A new model for Portuguese SRL.
    \item A comparison of the performance of multilingual and monolingual models.
    \item We show that using data from a high-resource language improves the scores of multilingual models in a low-resource language.
    \item We show that pre-training with dependency parsing can help models identify argument span boundaries when we have unclean text.
    \item We share our code and models trained with the complete set of available data (PropBank.Br + Buscapé).\footnote{\url{https://github.com/asofiaoliveira/srl_bert_pt}}.
\end{itemize}

\section{Related Work}
\label{related_work}

One of the most prominent early works in automatic Semantic Role Labeling was by Gildea and Jurafsky \cite{gildea_and_jurafsky}. They trained a system based on statistical classifiers on a part of the FrameNet dataset \cite{framenet}.
Since then, several models have been proposed, based first on machine learning algorithms and later on different neural network architectures. Currently, the models with the best performance in English SRL include in their architecture a pre-trained Transformer-based model. Examples of this are the works of Shi and Lin \cite{sota} and Li et al. \cite{structured}.

In Portuguese, there have been fewer models proposed, most of them based on techniques proven efficient in English.
Alva-Manchego and Rosa \cite{srl_pt1} used machine learning algorithms with features extracted from the gold parse trees of PropBank.Br v1 to predict arguments. 
Fonseca and Rosa \cite{srl_pt2} built a system similar to that of Collobert et al. \cite{collobert}, with no parsing information, but divided the task into two steps -- argument identification and classification.  
Hartmann et al. \cite{hartmann} compared the two previous systems using PropBank.Br versions 1.1 and 2 and automatically parsed trees. They concluded that the model of Alva-Manchego and Rosa performed better.

The most recent model we know for supervised Portuguese SRL is by Falci et al. \cite{falci}. It follows the architecture of He et al. \cite{he} and is a 2-layer bi-LSTM using word embeddings and a global inference step with IOB and PropBank constraints.

\section{Data}
\label{data}

PropBank is a project to annotate text with semantic roles \cite{propbank2}. 
It divides roles into two types: numbered roles A0-A5 and non-numbered roles that represent modifiers or adjuncts, e.g., AM-TMP (temporal modifier) and AM-LOC (location modifier). The AM's are defined globally across predicates, whereas the numbered roles' meanings are specific to each predicate sense. They are defined in the predicate's frame file.

Besides the two mentioned types, there are also continuation roles and reference roles. Continuation roles (C-x, where x is any other role) are used when the argument is non-contiguous. Reference roles (R-x, where x is a non-continuation role) are used on pronouns that refer to other roles present in the sentence.

Propbank.Br\footnote{\url{http://www.nilc.icmc.usp.br/portlex/index.php/en/downloadsingl}} \cite{propbankbr} is the (Brazilian) Portuguese version of PropBank and it follows the same annotation style as the original, having a similar role set. Unlike the recent versions of PropBank, PropBank.Br only annotates arguments related to verbal predicates and is much smaller in size. 
There are two versions of this dataset from two different data sources. 
The first version is based on the Brazilian Portuguese portion of the Bosque corpus, which is a part of the treebank ``Floresta Sintá(c)tica" \cite{floresta_sintactica}. It contains sentences extracted from the newspaper ``Folha de São Paulo" of 1994 which have been manually annotated. 

The second version of the dataset contains sentences extracted from the PLN-Br corpus \cite{plnbr}, also a journalistic corpus based on ``Folha de São Paulo". The sentences in this version were automatically parsed by PALAVRAS \cite{palavras}. The annotation project of this second version produced also a smaller annotated corpus, named Buscapé, which is based on a corpus of product reviews \cite{buscape}. Buscapé contains many misspelled words, poorly constructed sentences, verbs conjugated in the wrong tense and accents missing or in excess. This is very common in texts written on the internet in more casual settings. Thus, it represents an important and realistic benchmark for model testing.
The frame files for PropBank.Br are called Verbo-Brasil\footnote{Available at \url{http://143.107.183.175:12680/verbobrasil/sobre.php?lang=eng}} \cite{verbo_brasil,verbo_brasil2} and were based on PropBank's frame files.

In some experiments, we also use the English CoNLL-2012 SRL data \cite{conll2012_robust} and the Universal Dependencies\footnote{\url{https://universaldependencies.org}} (UD) Portuguese data set, UD-Portuguese\_Bosque \cite{ud}
. The former is a data set in English for the SRL task which had to be pre-processed to match the Portuguese data we use in our work. The latter is a dependency parsing data set.

\section{Model and Evaluation Methodology}

\begin{figure*}[!htb]
    \centering
    \includegraphics[width=0.8\linewidth]{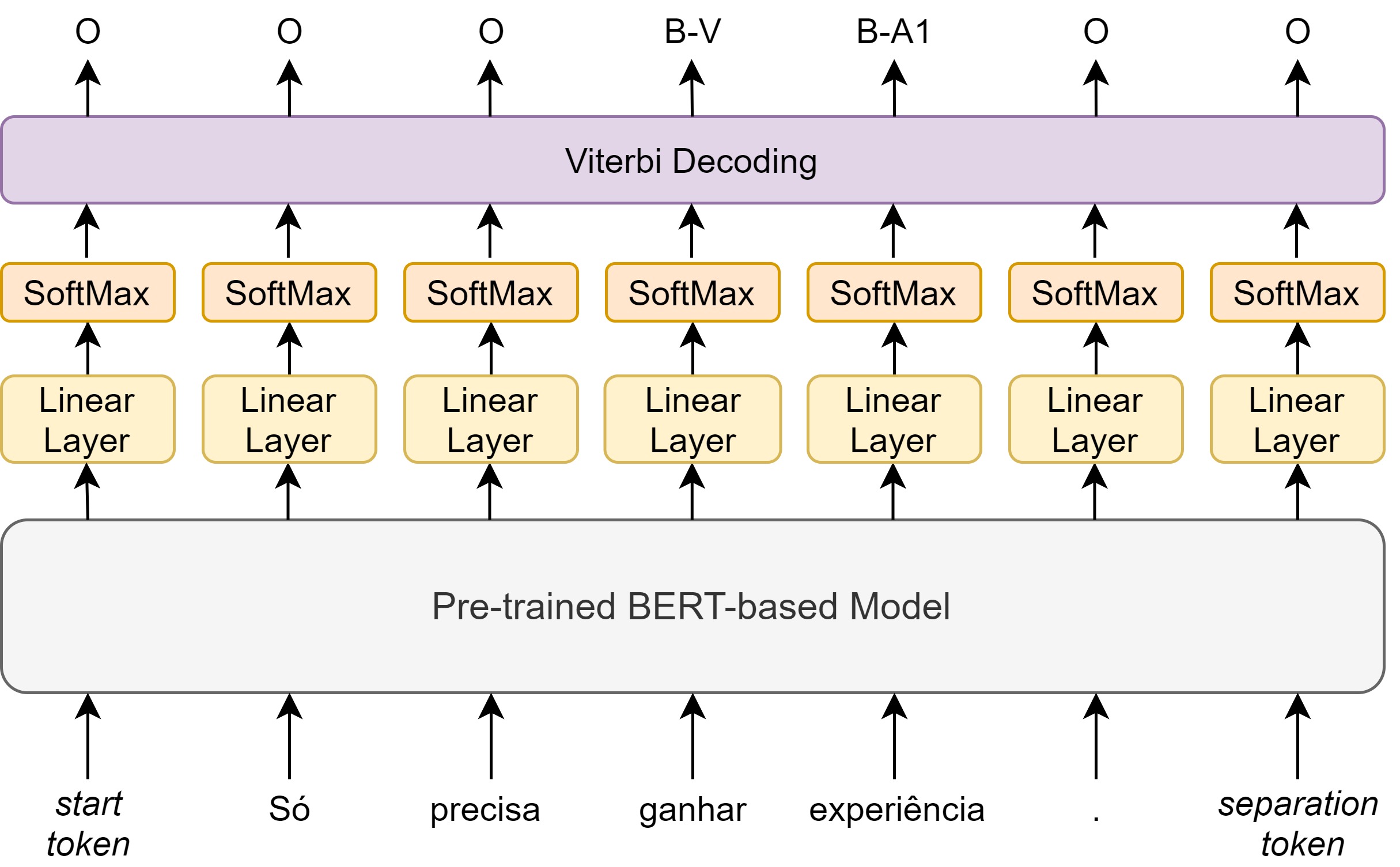}
    \caption{Model architecture with an example for the verb ``ganhar" (to gain) in the sentence ``Só precisa ganhar experiência" ([Someone] Just needs to gain experience).}
    \label{fig:architecture}
\end{figure*}

The model architecture used in our experiments is presented in Figure \ref{fig:architecture}. It includes a pre-trained transformer-based model on top of which a linear layer, a softmax function and Viterbi decoding are applied. The linear layer's parameters are randomly initialized and both these and the pre-trained model's parameters are tuned during training.
The Viterbi algorithm is conditioned on
the output having to be a valid IOB sequence, i.e., ``I-x" tags must be preceded either by ``B-x" or ``I-x", where x is any semantic role label. 

Our architecture performs neither predicate identification nor predicate disambiguation. Assuming we know the verbs' positions in a sentence (included in the datasets for training and testing), the sentence is passed to the SRL architecture one instance at a time. We call instance to the set of the sentence's tokens and a specific verb. The verb's position is indicated through the segment embeddings input (which are added to the token embeddings and the position embeddings \cite{bert}). This architecture does not use predicate sense disambiguation to make predictions. The identification of arguments and their classification with the appropriate semantic role are done at the same time.

In the rest of this section, we first describe the methodology for comparing our approach with the state of the art for Portuguese (Section \ref{outline}). Then we propose a methodology of our own to compare our architecture in different training conditions (Section \ref{method}).

\subsection{Methodology for comparison with Portuguese state of the art}
\label{outline}

We compare our proposed approach using the Portuguese pre-trained transformer-based models with the model of Falci et al. \cite{falci} (our baseline). We first pre-process the PropBank.Br v1.1 CoNLL-format dataset using the same steps as the baseline, and then perform 20-fold cross-validation with the same folds. A brief description of the dataset for this comparison is presented in Table \ref{tab:ds} (column ``Met.1").
After training, the models were evaluated with the script provided in the CoNLL-2005 Shared Task \textit{srl-eval.pl}\footnote{Available at \url{https://www.cs.upc.edu/\string~srlconll/soft.html}}.

\subsection{Methodology for comparing our models}
\label{method}

We will compare models trained in the Portuguese data (but based on different pre-trained transformer-based models) with models trained in both English and Portuguese and with models pre-trained first on dependency parsing.

To this end, we perform 10-fold cross-validation. We use stratified sampling for multi-label data \cite{stratification} to create the folds from the complete PropBank.Br dataset (versions 1.1 and 2), using the \textit{iterative-stratification}\footnote{Available at \url{https://github.com/trent-b/iterative-stratification}} package. Each fold is used as a test set once. The remaining folds are again sampled to obtain a validation set of approximately the same size of the test set. In addition to these, the out-of-domain Buscapé corpus was used to test all model runs.

The XML files of the three datasets (PropBank.Br v1.1, PropBank.Br v2 and Buscapé) were all pre-processed in the same fashion. We eliminated instances with more than one label for a word, we separated expressions joined with ``\_" and re-joined words formed by prepositional contractions. Additionally, we removed arguments labeled as ``AM-MED" or ``AM-PIN" because there is no mention of these labels in the annotation guides and for the corpora PropBank.Br v2 and Buscapé, we removed any instances with flags ``WRONGSUBCORPUS", ``LATER" or ``REEXAMINE", since, according to the guide, these indicate something wrong with the sentence that prevents its annotation.

In contrast to previous works, we only annotated an argument with a ``C-" role if the respective words were non-contiguous to the original argument. The CoNLL file for PropBank v1.1 separates arguments if they are not in the same node in the constituency-based syntactic tree, even if they are contiguous.

As in the previous methodology, the models were evaluated with the official script from the CoNLL 2005 Shared Task. Table \ref{tab:ds} presents a brief description of both datasets in this methodology (columns ``Met.2 CV" and ``Met.2 Buscapé"), as well as of the CoNLL-2012 dataset.

\begin{table*}[!htb]
    \centering
    \caption{Description of the SRL datasets used in this work: the number of instances, the total number of arguments annotated and the number of different semantic role labels present in the dataset (excluding verb and continuation role labels).}
    \label{tab:ds}
    \begin{tabular}{|l|c|c|c|c|}\hline
         & Met.1 & Met.2 CV & Met.2 Buscapé & CoNLL-2012 \\\hline\hline
        \textbf{\#Instances} & 5600 & 13665&709 & 316155\\\hline
        \textbf{\#Annotated Args}& 13135& 30459 & 1364 & 659297 \\\hline
        \textbf{\#Semantic Roles} & 18 & 26 & 23 & 21\\\hline
    \end{tabular}
\end{table*}

\section{Experiments}
\label{experiments}

We will now run experiments that try to answer the following research questions:
\begin{enumerate}
    \item Do new developments in models for semantic role labeling in English bring improvements to the task in Portuguese? 
    \item How do the state-of-the-art multilingual language models compare to existing monolingual models for Portuguese SRL? 
    \item Does cross-lingual transfer learning from English help the multilingual models' performance in Portuguese SRL? 
    \item Can we improve the results of the SRL task by training the language model on dependency parsing previously? 
    \item Is the Portuguese data useful or can we rely on models trained with English only? 
\end{enumerate}

The advantage of multilingual models is that they can learn the task in other languages and boost performance in Portuguese. Compared to English, Portuguese has much less SRL annotated data. Li et al. \cite{structured} showed that less data leads to poorer performance of models, even with powerful models as the ones being tested -- a drop from 86.47$F_1$ to 75.96$F_1$ with the RoBERTa model when using only 3\% of the CoNLL-2012 data. Since the annotation of more data is expensive, one way to attempt to mitigate this impact is to train a SRL multilingual model in other languages and use the trained model parameters as a starting point for training in Portuguese. This type of cross-lingual transfer learning, where information from high-resource languages is used in a low-resource language, has already been proven useful for other tasks \cite{choosing_lang}. 

On the other hand, syntax has often been used to boost SRL performance in proposed architectures. It is therefore relevant to see if it can also help Transformer-based architectures, particularly in low-resource languages. We integrate syntax using intermediate-task fine-tuning, which has at times led to improvements in models' scores for other tasks \cite{transfer_learning}.

\section{Results and Analysis}
\label{results}

The models were implemented in Python using the package \textit{AllenNLP} v1.0.0rc3 \cite{allennlp}, the (cased) mBERT, XLM-R$_\text{base}$ and XLM-R$_\text{large}$ models trained by \textit{Hugginface Transformers} v2.9.0 \cite{transformers_package} and the Portuguese BERT models (base and large), which will be referred to as brBERT, trained by \textit{neuralmind-ai} \cite{bertpt} and built on PyTorch v1.5.0 \cite{pytorch}. 

In each experiment, all models are trained with the same hyperparameters. With the exception of number of epochs, batch size and learning rate, all hyperparameters used followed the defaults in \textit{AllenNLP}'s English BERT SRL training configuration\footnote{\url{https://github.com/allenai/allennlp-models/blob/v1.0.0rc3/training_config/syntax/bert_base_srl.jsonnet}}.

\subsection{Comparison with Portuguese state of the art}
\label{sota}

As mentioned, work in Portuguese SRL has been scarce. From the models mentioned in Section \ref{related_work}, most are trained only on PropBank v1.1 and achieve less than 70$F_1$ in cross-validation. The exception is Alva-Manchego's model, trained in PropBank v2 and tested in PropBank v1.1 by Hartmann et al. \cite{hartmann}.
However, we compare only to the work of Falci et al. \cite{falci}, since they describe their methodology in detail, allowing us to compare the models trained in the exact same data.

We ran both brBERT$_{\text{base}}$ and brBERT$_{\text{large}}$ on the PropBank.Br v1.1 corpus pre-processed as described in section \ref{outline} with a batch size of 16 and a learning rate of $4\times10^{-5}$ for up to 100 epochs with early stopping after 10 epochs without improvement. The parameters are based on \textit{AllenNLP}'s model configuration, but due to memory constraints, the batch size had to be reduced (from 32 to 16); the learning rate was reduced as well (from $5\times10^{-5}$ to $4\times10^{-5}$) due to the fact that some folds' models were overfitting on no argument predictions.
In Table \ref{tab:results1}, we present the results for the best model out of all folds and the average $F_1$ across folds\footnote{Note that throughout the text, whenever average results are mentioned, we mean averaged across folds.}, since these were the metrics reported by our baseline. The values for the baseline are taken from Falci et al. \cite{falci}.

\begin{table}
    \centering
    \caption{Comparison of our proposed BERT-based model with the bi-LSTM-based baseline on 20-fold CV. The results for the baseline were taken from Falci et al. \cite{falci}.}
    \label{tab:results1}
    \begin{tabular}{|c|c|c|c|c|}
    \hline
        \multirow{2}{*}{\textbf{Model}} & \multicolumn{3}{| c |}{\textbf{Best Model}} & \multirow{2}{*}{\textbf{Average $F_1$}} \\
     \cline{2-4}
     & $P$ (\%) & $R$ (\%) & \textbf{$F_1$} &\\ \hline \hline 
        Baseline & 67.62 & 68.75 & 68.18 & 65.63 \\ \hline
        brBERT$_{\text{base}}$ & 84.24 & 85.37 & 84.80 & 81.48 \\ \hline
        brBERT$_{\text{large}}$  & 85.98 & 84.53 & 85.25 & 82.54 \\ \hline
    \end{tabular}
\end{table}

The BERT models give better results than the bi-LSTM model of the baseline. The improvements on the average $F_1$ are of 15.85 $F_1$ and 16.91 $F_1$ for the brBERT$_{\text{base}}$ and brBERT$_{\text{large}}$ models, respectively. This was expected due to the superior performance of BERT models in the English language. The difference is larger than the one observed in English, however, likely due to the fewer resources used in Portuguese, which hindered the LSTM-based model.

\subsection{Comparing monolingual to multilingual models}
\label{comparison}

In this section, we perform a more robust evaluation of the different pre-trained Transformer-based models by following a different methodology, described in Section \ref{method}, which uses more data and different pre-processing steps. We also experiment with alternative techniques for further improvement, particularly cross-lingual transfer learning from English.

We ran all of the models in this section with a batch size of 4 (due to memory constraints) and a learning rate of $1\times10^{-5}$ for up to 100 epochs with an early stopping after 10 epochs without improvement. The proposed architecture with the different pre-trained models was run in three scenarios:
\begin{enumerate}
    \item Fine-tuning only with Portuguese data; 
    \item Fine-tuning first with pre-processed CoNLL-2012 data for 5 epochs, followed by fine-tuning with Portuguese data (only for multilingual models, represented by a superscript ``\texttt{+}En", e.g., XLM-R$_{\text{base}}^{\texttt{+}\text{En}}$).
    \item Intermediate task fine-tuning in dependency parsing for 10 epochs, before fine-tuning in SRL (for the best models from the previous scenarios, represented by a superscript ``\texttt+UD").
\end{enumerate}

\subsubsection{Overall Results}
\label{overall}

Table \ref{tab:results3} presents the average precision, recall and $F_1$ measure of the 10 models (one for each fold) in their respective test sets and in the Buscapé set.

\begin{table*}[!htb]
    \centering
    \caption{Average of results of each model in the test set and Buscapé set.}
    \label{tab:results3}
    \begin{tabular}{|c||c|c|c|c||c|c|c|c|}
    \hline
        \multirow{2}{*}{\textbf{Model}} & \multicolumn{4}{| c ||}{\textbf{Average of Test Folds}} & \multicolumn{4}{| c |}{\textbf{Average of Buscapé}} \\
     \cline{2-9}
     & $p$ (\%) & $r$ (\%) & \textbf{$F_1$} & $\delta F_1$ &$p$ (\%) & $r$ (\%) & \textbf{$F_1$} & \textbf{$\delta F_1$} \\ \hline \hline 
        brBERT$_{\text{base}}$  & 75.78 & 76.83 & 76.30 &  & 74.00 & 72.68 & 73.33 &\\ \hline
        brBERT$_{\text{large}}$  & 76.65 & 78.20 & 77.42 & & 75.58 & 74.14 & 74.85 & \\ \hline
        XLM-R$_{\text{base}}$ & 74.42 & 76.04 & 75.22 &  & 73.12 & 72.54 & 72.82 &\\ \hline
        XLM-R$_{\text{large}}$ & 76.74 & 78.47 & 77.59 &  & 74.36 & 73.34 & 73.84 &\\ \hline
        mBERT & 72.34 & 73.20 & 72.76 &  & 67.10 & 66.70 & 66.89 &  \\\hline
        \multicolumn{9}{|c|}{\textit{Cross-lingual Transfer Learning}}\\ \hline
        XLM-R$_{\text{base}}^{\texttt+\text{En}}$ & 76.09 & 76.93 & 76.50 & 1.28 & 74.29 & 73.22 & 73.74 & 0.92 \\ \hline
        XLM-R$_{\text{large}}^{\texttt+\text{En}}$ & \textbf{77.71} & \textbf{78.75} & \textbf{78.22} & 0.63 & 75.36 & 73.77 & 74.55 & 0.71 \\ \hline
        mBERT$^{\texttt+\text{En}}$ & 74.22 & 75.56 & 74.88 & 2.12 & 69.41 & 68.98 & 69.19 & 2.3 \\ \hline
        \multicolumn{9}{|c|}{\textit{Transfer Learning from Syntax}}\\ \hline
        brBERT$_{\text{large}}^{\texttt+\text{UD}}$  & 76.90 & 78.19 & 77.53 & 0.11 & 75.25 & 73.75 & 74.49 & -0.36 \\ \hline
        XLM-R$_{\text{large}}^{\texttt+\text{UD}}$ & 77.00 & 78.40 & 77.69 & 0.10 & 75.77 & 74.08 & 74.91 & 1.07 \\     \hline
        XLM-R$_{\text{large}}^{\texttt+\text{En}\texttt+\text{UD}}$ & 77.38 & 78.57 & 77.97& -0.25 & \textbf{75.69} & \textbf{74.44} & \textbf{75.05} & 0.50 \\ \hline
    \end{tabular}
\end{table*} 

Note that the values of the monolingual brBERT models are smaller than the ones reported in the previous section because those values were from the validation set and these are from test sets. Moreover, the results in this section are evaluated in more data, with more challenging roles (scarcer during training) and include continuation arguments, which were removed for the previous section.

As expected, large models have a superior performance with respect to their base counterparts. XLM-R performs better than mBERT, which had already been reported for other tasks (\cite{xlmr}). Moreover, all models have a drop in all measures in the out-of-domain Buscapé set when compared to the test folds of the PropBank dataset. This is likely due to this dataset being more difficult, as previously discussed.

\subsubsection{Argument Identification and Classification}

The SRL task can be seen as two sub-tasks: argument identification and argument classification.  The performance in the classification, and, therefore, the scores obtained in SRL, are constrained by argument identification, since only correctly identified spans can be checked for the correct label.

\begin{figure*}[!htb]
    \centering
    \includegraphics[width=0.8\linewidth]{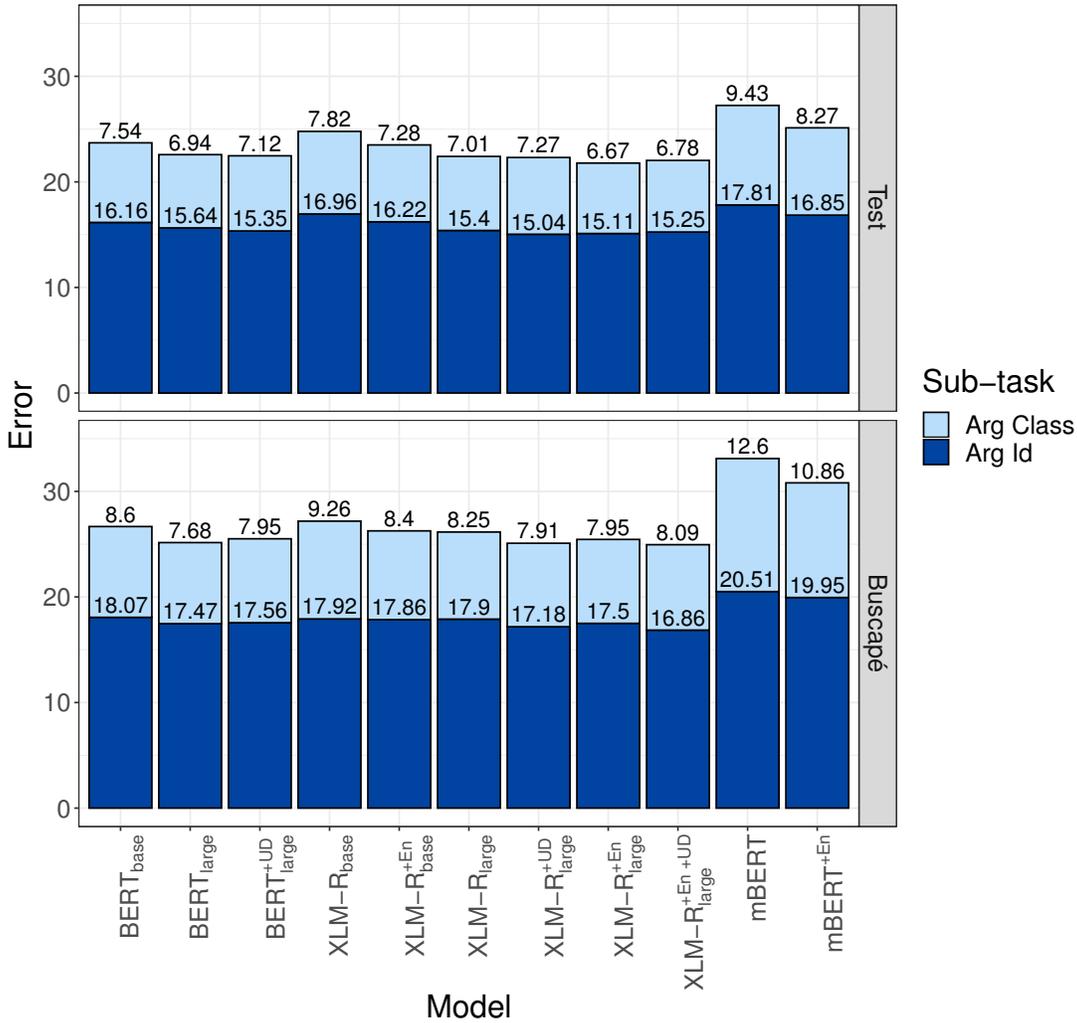}
    \caption{Average error across folds from the argument identification (``Arg Id") and argument classification (``Arg Class") for each model.}
    \label{fig:errors}
\end{figure*}

Additionally, the total error of the model can be decomposed in the error in identifying argument spans and the error in classifying the correctly identified spans.
In Figure \ref{fig:errors}, we report the average error for each model in each dataset decomposed into error from argument identification (``Arg Id") and from argument classification (``Arg Class").

Several things are noticeable in this figure. To start with, the error from the non-identification of arguments is much larger than the error from the mislabeling of arguments for all models and both datasets. Hence, our models are better at attributing semantic roles than at identifying argument spans.
Secondly, errors are higher in Buscapé. We cannot blame the drop in performance on one of these sub-tasks -- both are worse.
Lastly, the difference in $F_1$ between models is due both to differences in the identification and the classification of arguments.

\subsubsection{How do the state of the art multilingual language models compare to existing monolingual models for Portuguese SRL?} When training only with Portuguese data, brBERT$_{\text{large}}$ and XLM-R$_{\text{large}}$ have similar scores in the test sets, but the monolingual model outperforms the multilingual by approximately 1$F_1$ in the out-of-domain dataset. We believe that the monolingual model, having been pre-trained only in Portuguese data, has learned a better language structure, compared to XLM-R, and, therefore, can better understand the Buscapé data, despite the errors it contains. As for the base models, there is a larger difference between brBERT$_{\text{base}}$ and XLM-R$_{\text{base}}$ in the test set than in the Buscapé set, but the difference is still relatively small (approximately 1$F_1$). The mBERT model performs the worst: over 3$F_1$ points below brBERT$_{\text{base}}$.

\subsubsection{Does cross-lingual transfer learning from English help the multilingual models’ performance in Portuguese SRL?} When including cross-lingual transfer learning, all multilingual models get an increase in their scores. The less powerful the model, the larger is this increase.
In this scenario, XLM-R$_{\text{large}}$ achieves the best score in the average of the test folds, but is still slightly behind brBERT$_{\text{large}}$ in Buscapé. On the other hand, XLM-R$_{\text{base}}$ improves enough to become on par with brBERT$_{\text{base}}$ in both datasets. As before, mBERT underperforms compared to all other models.

\subsubsection{Can we improve the results of the SRL task by training the language model on dependency parsing first?}
\label{transfer_learning}

Using the UD dataset did not have a positive impact in the performance of Portuguese SRL for the monolingual BERT$_\text{large}$ model. For the multilingual XLM-R$_{\text{large}}$ and XLM-R$_{\text{large}}^{\texttt+\text{En}}$ models, fine-tuning first with UD boosts the performance in the Buscapé dataset. In fact, with the UD data, we improve the best results obtained before on Buscapé.

Additionally, both models that were fine-tuned in UD are in general better at argument identification in Buscapé than BERT$_{\text{large}}$, leading us to believe that the use of syntax may help identify spans in more difficult data. This agrees with previous work, such as Strubell et al. \cite{strubell}, which found that syntax helped with span boundary identification.
However, the differences are small, so we cannot be certain such differences are not just accidental and based on the little test data available for this task.

\subsubsection{Is it useful at all to use the Portuguese data or can we rely on models trained with English data only?}
\label{zero_shot}

We provide in Table \ref{tab:zero_shot} the zero-shot transfer learning results from the three multilingual models trained in the CoNLL-2012 data, i.e., the models were fine-tuned in English and tested in Portuguese.

\begin{table}[!htb]
    \centering
    \caption{Results for zero-shot cross-lingual transfer learning. The three models were trained on the pre-processed CoNLL-2012. The results are the average of the 10 folds and the result in the Buscapé corpus.}
    \label{tab:zero_shot}
    \begin{tabular}{|c||c|c|c||c|c|c|}\hline
    \multirow{2}{*}{\textbf{Model}} & \multicolumn{3}{| c ||}{\textbf{Average of Test Folds}} & \multicolumn{3}{| c |}{\textbf{Buscapé}} \\
     \cline{2-7}
     & \textbf{$p$} (\%) & \textbf{$r$} (\%) & \textbf{$F_1$} & \textbf{$p$} (\%) & \textbf{$r$} $(\%)$ & \textbf{$F_1$}\\ \hline \hline 
        mBERT & 60.73 & 65.59 & 63.07 & 57.83 & 59.31 & 58.56 \\ \hline
        XLM-R$_{\text{base}}$ & 63.58 & 69.90 & 66.59 & 63.56 & 67.01 & 65.24 \\ \hline
        XLM-R$_{\text{large}}$ & 64.64 & 70.85 & 67.60 & 63.05 & 66.94 & 64.94 \\ \hline
    \end{tabular}
\end{table}

Despite the large drop in $F_1$ measure between zero-shot cross-lingual transfer learning and training only with Portuguese data -- between 9 and 10 $F_1$ points --, it is encouraging to see that the former still perform reasonably well in the Portuguese data. It means languages with no annotated SRL data can use multilingual models trained for this task in English and obtain acceptable results. However, the gains of having own language resources are evident.

\subsubsection{Choosing the Best Model}
\label{choosing}

\begin{figure*}[!htb]
    \centering
    \includegraphics[width=0.8\linewidth]{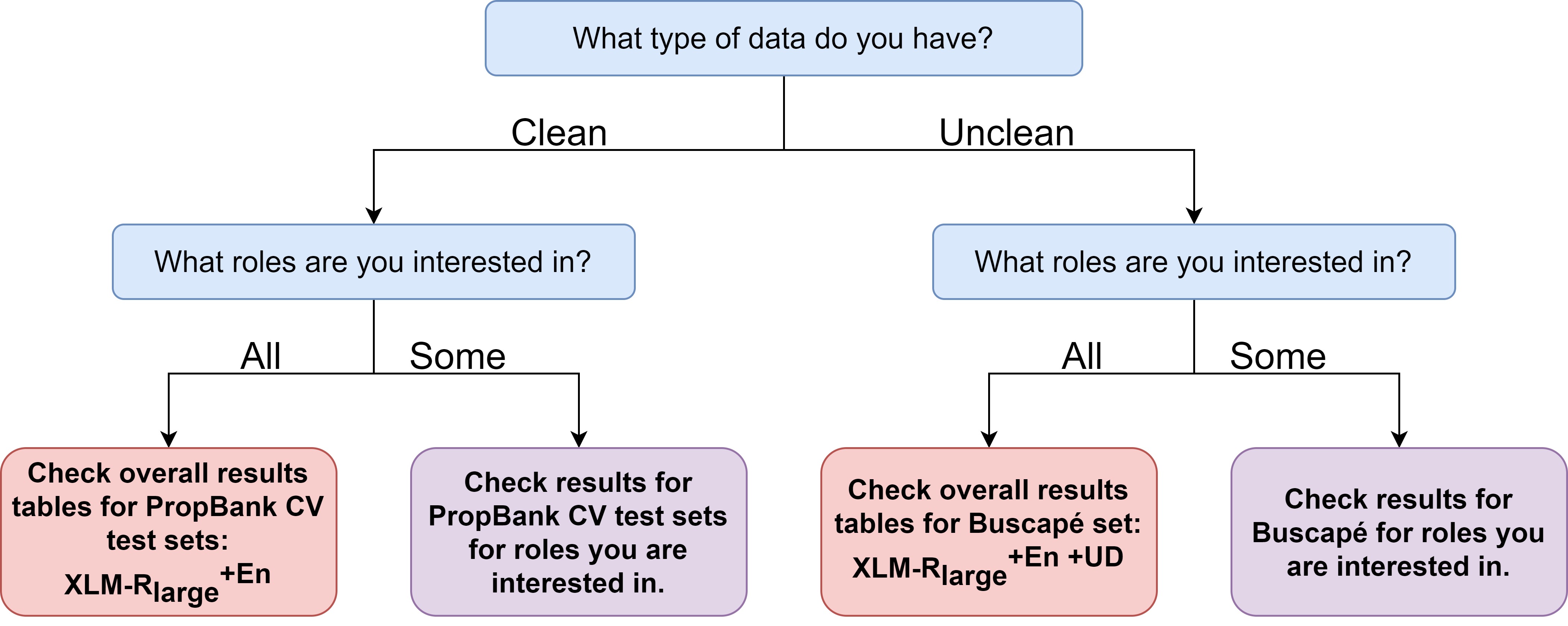}
    \caption{Heuristic for choosing the most appropriate model in different situations.}
    \label{fig:dec_diag}
\end{figure*}

All things considered, choosing the best model is not an easy task. It depends on many factors regarding the intended application. In this section, we provide an heuristic for choosing a model for an application, based on the obtained results (Figure \ref{fig:dec_diag}). 

First of all, it is important to determine the type of data for both training and inference. If the model is to be applied to text of a more formal variety, where there is some certainty that sentences will be properly structured and words properly spelled (``clean" data, e.g. journalistic text), it is better to consider the performance achieved in the PropBank CV test sets. On the other hand, if one is dealing with text from online sources, where there is no guarantee of the mentioned requirements, or it is certain they are not met (``unclean" data), it is better to look at the scores from the Buscapé set. 

The overall scores of the models may not be the most appropriate to distinguish between them in all situations. However, the distribution of roles in these datasets is likely to be representative of that found in the Portuguese language. Therefore, if one is interested in the best model for the language in general, they can choose the one with the highest $F_1$ in the relevant dataset. 
If, on the other hand, one is interested only in a subset of roles, it is best to choose the best model for those roles according to the results per roles (an implementation of this heuristic is available with the code for this article).

The results obtained allow us to compare the models, as intended. However, these scores may not correspond to the model's actual performance for two reasons. There are semantic roles that were only annotated in the second version of PropBank, therefore the results for these are not reliable, as the data is inconsistent. 
On the other hand, upon inspecting the resulting predictions and gold labels, there seem to be some issues with the annotated sentences. For example, when predicting the arguments for the verb ``retirar" (withdraw) in the sentence ``Os EUA devem retirar suas tropas da Somália até março." (The USA should remove their troops from Somalia until March.), the gold labels indicate that ``suas tropas da Somália" (their troops from Somalia) corresponds to A1 -- ``entidade ou coisa sendo retirada" (entity or thing being removed). However, we believe that A1 should be ``suas tropas" (their troops) and A2 -- ``local de onde foram retiradas" (place from where they were removed) -- should be ``da Somália" (from Somalia)\footnote{The sentence could also be translated as ``The USA should withdraw their Somalian troops until March", and this ambiguity is likely where the confusion comes from, but this sentence makes less sense.}. We believe this is due to the usage of automatically annotated parse trees in version 2, which constrained the SRL annotations -- in our example, ``suas tropas" did not appear as a constituent in the tree, therefore the annotator could not properly classify it. Nonetheless, this dataset is extremely useful in such a low-resource language.
We mention this merely as a caveat to warn possible users not to take the obtained scores as fully representative of the model's performance.

\section{Conclusions}

Our architecture achieves a new state of the art for SRL in Portuguese, improving previous results by over 15$F_1$ points. 
Interestingly, we found that, with the techniques employed, the multilingual model XLM-R$_\text{large}$ could achieve better results than the monolingual BERT$_\text{large}$
. This suggests that monolingual models may become unnecessary when powerful multilingual models are available, at least for low-resource languages, such as Portuguese. 

We found that the largest percentage of errors in all models comes from the non-identification of arguments, instead of their misclassification.
Thus, we suggest that future research focus on this sub-task. Our transfer learning experiment with dependency parsing helped in span identification. However, it is possible that constituency parsing would yield even better results, since arguments for a predicate are commonly constituents in the constituency parse tree. 
Another interesting possibility for future work would be to apply the constraints from Li et al. \cite{structured}, which were reported to help their RoBERTa model when trained with less data.
We consider the most important line of future work, however, to be the improvement of the Portuguese datasets, by harmonising the role set across versions and by having all of the datasets manually revised once again to eliminate any annotation errors that may exist.

\section*{Acknowledgments}

This work was developed in Project Text2Story, financed by the ERDF – European Regional Development Fund through the North Portugal Regional Operational Programme (NORTE 2020), under the PORTUGAL 2020 and by National Funds through the Portuguese funding agency, FCT - Fundação para a Ciência e a Tecnologia within project PTDC/CCI-COM/31857/2017 (NORTE-01-0145-FEDER-03185)

Daniel Loureiro is supported by the European Union and Fundação para a Ciência e Tecnologia through research grant DFA/BD/9028/2020



\bibliographystyle{IEEEtran}
%

\bibliography{IEEEabrv,bare_conf.bib}

\end{document}